\documentclass[conference]{IEEEtran}
\usepackage{times}
\usepackage{soul}
\usepackage{float}

\usepackage[numbers]{natbib}
\usepackage{multicol}
\usepackage[bookmarks=true]{hyperref}

\newcommand{\yash}[1]{}
\newcommand{\pranay}[1]{}
\newcommand{\charu}[1]{}
\newcommand{\kar}[1]{}
\newcommand{\chris}[1]{}
\newcommand{\rev}[1]{}

\newcommand\tent[1]{}

\newcommand\fixed[1]{}

\newcommand{\qc}[1]{Q_c}
\newcommand{\qa}[1]{Q_a}
\newcommand{\qt}[1]{Q_t}
\newcommand{\qr}[1]{Q_r}
\newcommand{\qq}[1]{Empir3D}

\newcommand{\cd}[1]{$D_{c}$}
\newcommand{\hd}[1]{$D_{h}$}
\newcommand{\emd}[1]{$D_{em}$}
\newcommand{\dcd}[1]{$D_{dcd}$}

\usepackage{graphicx} 
\usepackage{multirow}
\usepackage{multicol}
\usepackage{array,booktabs}
\usepackage[skins,theorems]{tcolorbox}
\usepackage{hyperref}
\usepackage{tabularx}
\usepackage{xcolor}
\usepackage[font=small,labelfont=bf]{caption}

\begin{document}

\title{\qq{} : A Framework for Multi-Dimensional Point Cloud Assessment}

\author{Yash Turkar, Pranay Meshram, Christo Aluckal, Charuvahan Adhivarahan, Karthik Dantu}



%

\let\oldtwocolumn\twocolumn
\renewcommand\twocolumn[1][]{%
    \oldtwocolumn[{#1}{
    \begin{center}
              \includegraphics[trim=1cm 1cm 1cm 1cm, clip=true,width=\textwidth]{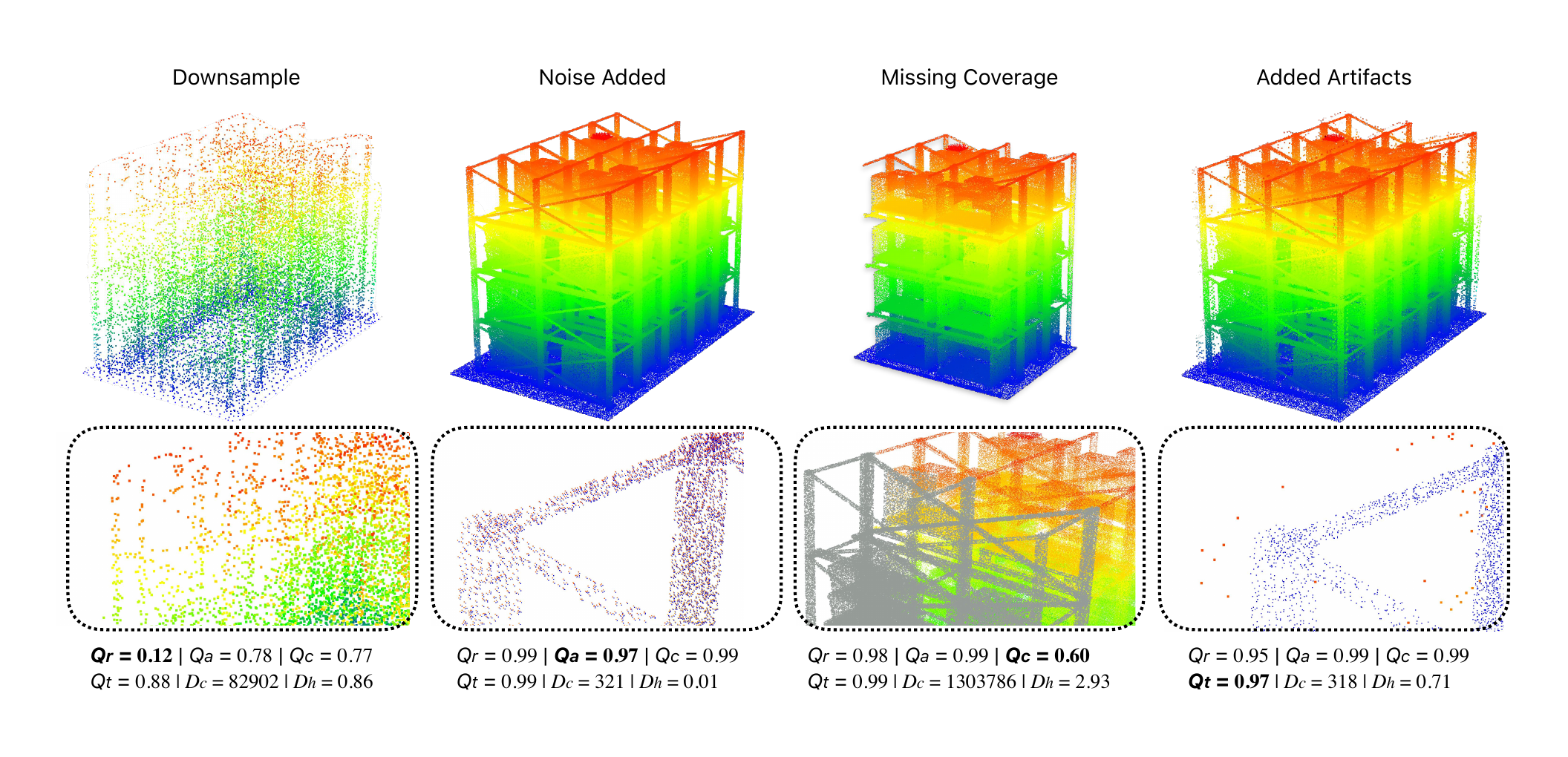}
              \captionof{figure}{
              Degradations on a sample point cloud (evaluated against original); Left to right: point cloud is, down-sampled, noise is added, cropped and artifacts added. Zoomed-in view below degraded point cloud, and the corresponding \qq{} metric reflects the degradation. \(Qr,\ Qa,\ Qc,\ Qt\) is resolution, accuracy, coverage, and artifact score respectively and \cd{} and \hd{} are Chamfer and Hausdorff distances}
              \label{intro_fig}
              \vspace{0.5cm}
        \end{center}
    }]
}

\maketitle

\begin{abstract}

Advancements in sensors, algorithms and compute hardware has made 3D perception feasible in real-time. Current methods to compare and evaluate quality of a 3D model such as Chamfer, Hausdorff and Earth-mover's distance are uni-dimensional and have limitations; including inability to capture coverage, local variations in density and error, and are significantly affected by outliers. In this paper, we propose an evaluation framework for point clouds (\qq{}) that consists of four metrics - resolution (\(\qr{}\)) to quantify ability to distinguish between the individual parts in the point cloud, accuracy (\(\qa{}\)) to measure registration error, coverage (\(\qc{}\)) to evaluate portion of missing data, and artifact-score (\(\qt{}\)) to characterize the presence of artifacts. Through detailed analysis, we demonstrate the complementary nature of each of these dimensions, and the improvement they provide compared to uni-dimensional measures highlighted above. Further, we demonstrate the utility of \qq{} by comparing our metric with the uni-dimensional metrics for two 3D perception applications (SLAM and point cloud completion). We believe that \qq{} advances our ability to reason between point clouds and helps better debug 3D perception applications by providing richer evaluation of their performance. 
Our implementation of \qq{}, custom real-world datasets, evaluation on learning methods, and detailed documentation on how to integrate the pipeline will be made available upon publication.\href{https://droneslab.github.io/E3D/}{{\color{blue}Project page}}.

\end{abstract}


\IEEEpeerreviewmaketitle

\section{Introduction}
\label{intro}
3D perception is crucial for various applications including autonomous driving~\cite{caesar2020nuscenes,oxfordrobot}, infrastructure inspection~\cite{bimlidar,VALENCA2017668}, augmented reality~\cite{mahmood2020bim, chen2019overview}, mobile manipulation~\cite{5354683,seita2023toolflownet}, 3D reconstruction~\cite{YI20171,brostow2008segmentation}, object detection~\cite{xuSqueezeSegV3SpatiallyAdaptiveConvolution2020,zhouVoxelNetEndtoEndLearning2017}, and GIS applications \cite{DBLP:journals/corr/abs-1908-08854, Westoby2012}. Each of these applications uses a 3D point cloud as input. Such 3D point clouds can be produced by various methods such as dense SLAM~\cite{xufastlio2fastdirect2022,Rosinol20icra-Kimera,labbe2019rtab}, structure-from-motion (Photogrammetry) \cite{Smith2015,Westoby2012,Schonberger_2016_CVPR}, survey grade scanners \cite{leicageosystemsRoboticTotal} and generative/learning-based methods \cite{yuan_pcn_2018, tchapmi_topnet_2019, pan_ecg_2020, Zhao_2021_ICCV, zhou2022seedformer, fan2017point, Wang_2018_ECCV}.

A logical question to be asked is how good a constructed 3D point cloud is in comparison to the real-world scene it represents and/or for the intended application. 3D point clouds are typically evaluated using distance-based similarity metrics comparing the constructed 3D point cloud with a reference. Such metrics quantify similarity of the unordered set of points in the constructed point cloud with the ones in the reference. Prevalent methods are  Chamfer distance (\cd{}), Hausdorff distance (\hd{}) \cite{Huttenlocher1990}, and earth-mover's distance (\emd{}) \cite{Fredman1987}.  Chamfer distance is the sum of squared distances between closest point pairs in two shapes. Earth-mover's distance is the sum of distances between closest point pairs where pairing is bijective and Hausdorff distance is the greatest of distances between closest point pairs. Though popular, each of these measures are uni-dimensional and have their own limitations in comparing two point clouds. \cd{} and \hd{} have limited sensitivity to point density and are significantly influenced by outliers~\cite{wu2021densityaware}. While \emd{} can detect changes in density, the bijectivity requirement can lead to ignoring local fine-grained structural details. Also, \emd{} is significantly more computationally expensive, which can limit its practicality. 

Some other relevant methods ~\cite{9857859,wu2021densityaware,1640800,DeepEMD} focus on specific applications like visual quality and point cloud generation acting as a loss function for neural network training.~\cite{1640800} provides a way to measure the accuracy and coverage of meshes generated by multi-view stereo reconstruction. 

Evaluating 3D point clouds is challenging, and requires quantification of multiple factors for a comprehensive comparison. Below are a list of factors that we deem important: 

\begin{itemize}
    \item It should reward a constructed point cloud if its density is high (resolution), it matches points from the reference (accuracy), and it is able to capture most of the reference points spatially (coverage)
    \item It should penalize the constructed point cloud if it has points in areas where the reference doesn't (artifacts)
    \item It should be computationally efficient to process large point clouds
\end{itemize}

To address these issues, we propose \qq{}, an \underline{E}valuation \underline{M}ethodology for \underline{P}o\underline{I}ntcloud \underline{R}easoning in \underline{3D}. \qq{} comprises of four metrics, each evaluating a specific aspect of point cloud quality:
\begin{itemize}
    \item \textbf{Resolution:} Resolution is the ability to resolve areas in a point cloud. It is an indicator of how detailed the point cloud is.
    \item \textbf{Accuracy:} Measures how close the points are to their true positions. 
    \item \textbf{Coverage:} Measures the areas of the reference that the constructed point cloud covers. Larger the score, larger the overlap between the two point clouds. 
    \item \textbf{Artifact Score:} Measures the proportion of anomalous points (artifacts) added in error in the constructed point cloud. 
\end{itemize}

To demonstrate these metrics visually, \autoref{intro_fig} provides a comparison between point clouds with some variations (down-sampling, adding noise, removing some portions, and adding artifacts). We show \qq{} metrics and two other popular point cloud comparison metrics - Chamfer distance (\cd{}) \cite{pmlr-v80-achlioptas18a} and Hausdorff distance (\hd{}) \cite{Huttenlocher1990} 

Design of the \qq{} metrics took multiple iterations striving to maximize two aspects - comprehensive evaluation of point cloud quality while ensuring independence across metrics to limit the number of metrics we used to capture the comparison. As a result, \qq{} provides a detailed evaluation of point cloud quality by addressing various aspects of point clouds, making it a valuable tool for several applications. The contributions of this paper are as follows:
\begin{itemize}
    \item We propose \qq{}, a comprehensive framework for evaluating constructed point clouds with a reference.
    \item We evaluate the framework on a real-world, simulated and generated point clouds demonstrating the utility of each metric for real-world applications.
    \item Perform an ablation study to show changes in metrics caused by a change to point clouds.
\end{itemize}

\section{Related Work}
\label{related}


New-age sensors like high-resolution cameras, LiDARs, and RADARs can produce rich, dense point clouds. Visual SLAM systems that use monocular cameras \cite{murartal17}, stereo cameras \cite{labbe14}, and RGB-D cameras \cite{endres14} to produce dense \cite{endres14,labbe14} or sparse \cite{murartal17} point clouds have been proposed. Photogrammetry tools like \cite{schoenberger2016mvs,schoenberger2016sfm,moulon2016openmvg} have enabled quick and easy access to reconstructing 3D environments for applications in game development, augmented reality and geospatial surveying.



\subsection{Point clouds from 3D reconstruction}

Recent advances in sensor technology, efficient libraries~\cite{Agarwal_Ceres_Solver_2022,Rusu_ICRA2011_PCL}, Neural-network architectures \cite{qi_pointnet_2017} and faster computing have enabled real-time dense mapping. Subsequently, Vision and LiDAR based SLAM systems~\cite{zhangLOAMLidarOdometry2014,Rosinol20icra-Kimera} have followed suit, especially in dense mapping performance. \cite{zhangLOAMLidarOdometry2014,shanLeGOLOAMLightweightGroundOptimized2018,shanLIOSAMTightlycoupledLidar2020,shanLVISAMTightlycoupledLidarVisualInertial2021,qinLINSLidarInertialState2020,xu_fast-lio2_2022,He2023} generate relatively dense point clouds using online localization and mapping. \cite{vizzoPoissonSurfaceReconstruction2021,zhong2023icra} output meshes by performing offline mapping and localization either solely with sequential LiDAR scans or with additional position information.

SLAM methods are generally evaluated for their localization and re-localization performance with the Absolute Trajectory Error (ATE) as seen in \cite{bujancaRobustSLAMSystems2021,6385773,6225199} with changes in environmental factors such as illumination. Although ATE is a good measure of a SLAM system's localization performance, it is a poor measure of map quality. In some cases, ATE can be used to evaluate the overall structure of the map, not density and completeness. For example, ORB-SLAM \cite{murartal17} is known for good localization and tracking performance, even though it produces sparse point cloud maps. In \cite{wisdom}, the authors use a WiFi-based distributed mapping system which cannot be evaluated with the ATE since a ground truth trajectory is hard to obtain in a distributed mapping scenario. Thus, the authors use known landmark (AprilTag~\cite{apriltags}) positions to evaluate their system, indicating a need for a metric to evaluate the map quality directly.
 ~\cite{6225199,6385773} highlights the lack of ground truth to evaluate point clouds; we address this by using simulated datasets as well as capturing ground truth using poses measured using a robotic total station and stitching corresponding LiDAR scans.

\subsection{Point clouds in Learning}


In addition to dense mapping and 3D reconstruction, contemporary learning methods and networks have enabled applications like object detection \cite{qiDeepHoughDetection2019, Zhou_2018_CVPR,Shi_2019_CVPR, Yang_2018_CVPR,Lang_2019_CVPR}, segmentation \cite{qiPointNetDeepHierarchical2017,Zhao_2021_ICCV,lai2022stratified,xu2020squeezesegv3} point cloud completion \cite{yuan_pcn_2018,tchapmi_topnet_2019,pan_ecg_2020,pan_variational_2021,Huang_2020_CVPR} , super-resolution \cite{dinesh3DPointCloud2019,guanFogHighResolutionImaging2020b,ledigPhotoRealisticSingleImage2017,shanSimulationbasedLidarSuperresolution2020,wuPointCloudSuper2019,zyrianovLearningGenerateRealistic2022} , image-to-point cloud generation\cite{fan2017point}, image-to-mesh generation \cite{Wang_2018_ECCV}, denoising and compression. These learning methods are usually evaluated using popular distance based metrics mentioned in \autoref{intro} and in some cases introduce non-standard metrics to capture subjective perceptual quality, which makes bench-marking a challenging task and highlights the need for a better metric.

\subsection{Point clouds in Multimedia and AR/VR}
Applications such as augmented and virtual reality, social media avatars and game development have greatly benefited from the recent rise in point cloud acquisition and processing techniques \cite{bonatto2016explorations}. These multimedia and AR/VR methods are generally evaluated for their perceptual quality and compression losses. \cite{yangInferringPointCloud2022,yangPredictingPerceptualQuality2021a,yangNoReferencePointCloud2022,liuPointCloudQuality2022,dinizColorGeometryTexture2021a,meynetPCQMFullReferenceQuality2020,violaColorBasedObjectiveQuality2020} propose no-reference and full-reference ways to evaluate the perceptual quality of point clouds with a focus on visual fidelity and predicting subjective quality. While these methods do a good job at evaluating the perceptual quality of the point clouds, they do not account for all the geometric aspects of quality.

\subsection{Popular distance-based metrics for point cloud comparison}

While distance based metrics can be convenient and relatively fast to compute, they are uni-dimensional and only output a single measure of similarity which cannot account of all aspects of quality. They also fail to reward and penalize candidate point clouds based on specific dimensions of quality making them difficult to be used as a reasonable feedback signal for improvement in learning methods and SLAM systems.

\subsubsection{Chamfer Distance (\cd{})}

\cd{} (\autoref{cd}) is computed as the sum of distances in two point clouds, usually referred to as source and candidate. For each point in the source, the distance to its nearest neighbor in the candidate point cloud is computed and vice versa. The sum of distances over both point clouds is the \cd{}. It is fast to compute, and it can capture the overall similarity between two point clouds. However, it does not account for the local variations and structural information in the point clouds, which can be important in some applications such as the ones mentioned in \autoref{intro}. Secondly, it is insensitive to density distribution and significantly influenced by outliers. While being influenced by outliers can provide insights into point cloud similarity, it can cause over-penalization where the candidate point cloud is unjustly penalized even if it has good coverage and accuracy.

\vspace{-.25em}
\begin{equation}
    D_{c}(A,B) =  \sum_{a\in A} \min_{b\in B} \lVert a-b \rVert_2 + \sum_{b\in B} \min_{a\in A} \lVert a-b \rVert_2
    \label{cd}
\end{equation}
\vspace{.25em}
\subsubsection{Hausdorff Distance (\hd{})}

\hd{} (\autoref{hd}) is calculated as the maximum distance between two points in the source and candidate point clouds. This means that for each point in one point cloud, the distance to the farthest point in the other point cloud is calculated, and the maximum of all such distances is the \hd{}. It captures the similarity between two point clouds, including their overall arrangement. However, it fails to capture any local variations and density of the point clouds.

\vspace{-.25em}
\begin{equation}
    D_{h}(A,B) = \max (\, \sup_{a\in A} \inf_{b\in B}d(a,b),\, \sup_{b\in B} \inf_{a\in A}d(a,b)\, )
    \label{hd}
\end{equation}
\vspace{.25em}

\subsubsection{Earth-mover's Distance (\emd{})}
\emd{} (\autoref{emd}) solves the optimal-transport problem, also known as the assignment or correspondence problem, by finding a bijective mapping between the two point sets. It is known that the optimal bijection is unique and is invariant to infinitesimal movement~\cite{fan_point_2016}.  While this makes \emd{} one of the most precise measures of distance-based similarity, its \(\sim O(n^2 \log n)\) \cite{Fredman1987} complexity makes it impractical for large point clouds. Additionally, the bijectivity requirement is not realistic when point clouds are in the \(10^7\) points range.

\begin{equation}
    D_{em}(A,B) = \min_{\phi:A\rightarrow B} \sum_{a\in A} \lVert a - \phi(a) \rVert_2
    \label{emd} 
\end{equation}
    where, \(\phi:A\rightarrow B \text{ is a bijection}\)

\section{\qq{} Framework}
\label{method}



\begin{figure}[htbp]
\vspace{-0.5cm}
\centering
\includegraphics[width=\columnwidth]{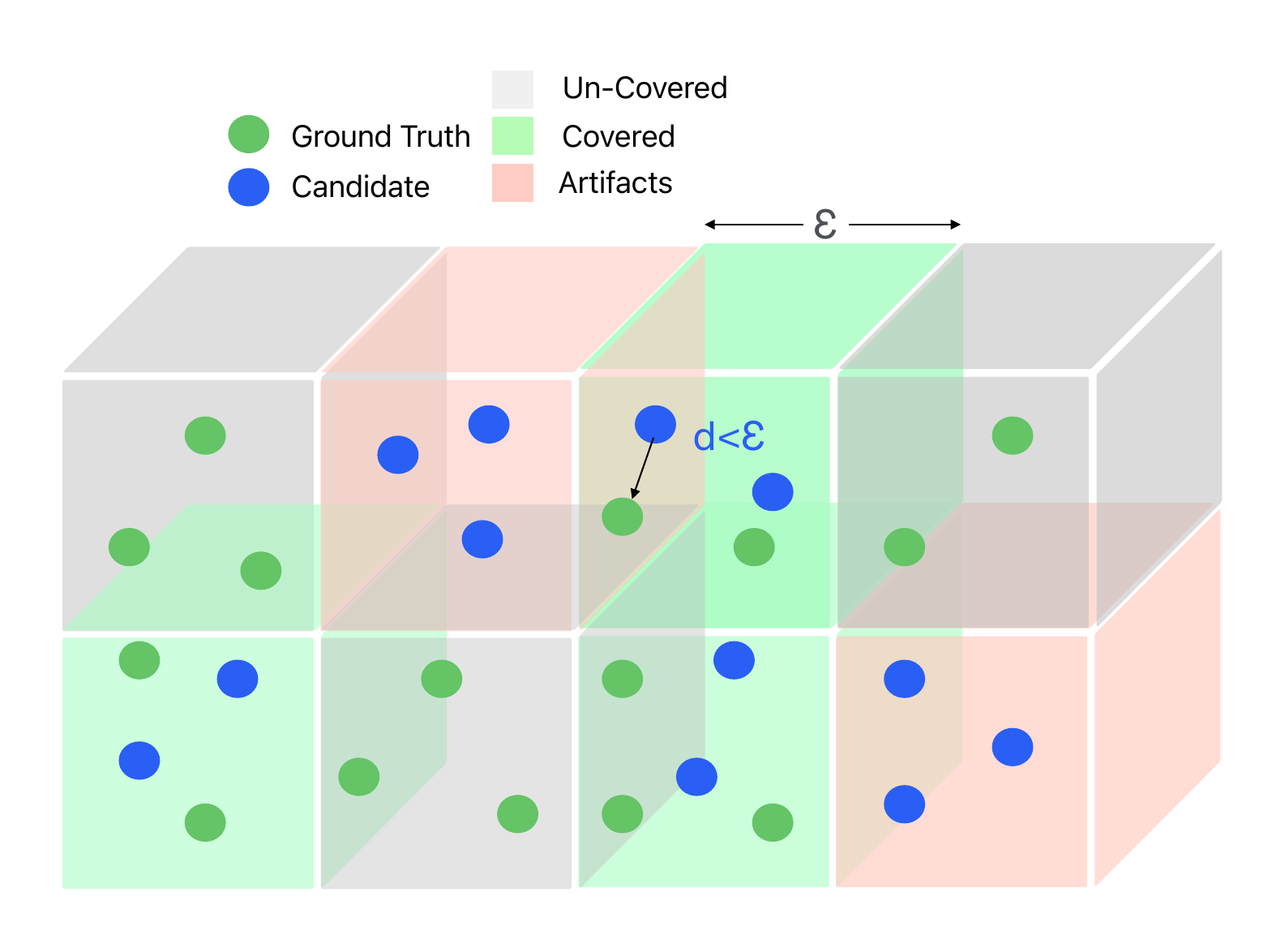}
\vspace{-0.8cm}
    \caption{Figure demonstrates cells of size \(\epsilon\), green cells contain both ground truth (green) and candidate (blue) points making them \emph{covered}, red cells only contain candidate points making them \emph{artifacts} and grey with only ground truth points showing \emph{missing coverage (or un-covered)}. Top row 3rd cell shows \(d<\epsilon\) as the distance considered to compute accuracy based on \autoref{accr}\label{pqm_animate}}
\end{figure}


The \qq{} framework provides a multi-dimensional comparison between two point clouds. This could be a ground truth point cloud and a captured or generated point cloud. We denote the source (ground truth) point cloud by $A={a_i}$, which we refer to as $pcd_A$. Similarly, we denote the candidate point cloud by $B={b_i}$, referred as $pcd_B$, where $a_i$ and $b_i$ are in $R^3$ and $i=1,\ldots,K$. Our goal is to measure the difference in quality between the source ($pcd_A$) and the candidate ($pcd_B$). 


The fundamental requirement of a high-quality point cloud is to represent the inherent continuous structure of a 3D environment or an object as best as possible. This is a challenging task because most sensors produce discrete outputs. Generative networks and mapping algorithms therefore either produce results in the form of point clouds or use interpolation to produce continuous meshes. The discrete nature of sensors, the error in their measurements, and the ability of the algorithm to handle these errors lead to variations in point cloud quality. Errors may also occur due to faulty depth estimation, low sample size that affects interpolation accuracy, poor generalization of neural-networks, and misplaced points due to errors in pose information when using SLAM.

We therefore define quality as a composition of four metrics: 
resolution, accuracy, coverage and artifact-score. Each metric contributes to the overall quality of the point cloud, and evaluating them independently enables us to assess the effect of each metric on the overall quality.
\(\qr{}\), \(\qa{}\), \(\qc{}\) and \(\qt{}\) denote the individual sub-metrics resolution, accuracy, coverage, artifact-score respectively.

\noindent {\bf Region Splitting}: 
To efficiently evaluate the point clouds, we divide them into smaller regions of equal size ($r$). This enables us to compute in parallel and provides insights into the values of the metrics of different areas within the point cloud. Point clouds are split into $N$ such regions, and metrics are computed for each. Regions are denoted as $reg_{A^{j}} \in pcd_A$ and $reg_{B^{j}} \in pcd_B$, where $j=1 \ldots N$.

\label{eps_def}
Independent of regions, we define \emph{cells} as volumes of size \(\epsilon\), where \(\epsilon\) is a hyper-parameter set by the user based on expected precision. Let set \(S = \{s_i | i = 1,2. \ldots M\}\) be a set of all cells such that the total volume occupied by S is equal to the total volume occupied by \(pcd_A\) and \(pcd_B\). Further, let \(S_A \subseteq S\) and \(S_B \subseteq S\) where \(S_A\) and \(S_B\) are sets of cells occupied by points of \(pcd_A\) and \(pcd_B\) respectively.

\noindent {\bf \qq{} Metrics}: 
Metrics are normalized between 0 and 1, representing the lowest and highest values of quality respectively. In contrast, geometric distance metrics such as \cd{}, \hd{}, and \emd{} are typically calculated such that a score of 0 represents a perfect match, and any value greater than 0 represents a degree of mismatch.


\subsection{Resolution \(\left(\qr{}\right)\)}
\label{Resolution_Def}
We define resolution (per region) as the ratio of the average distance between points of \(reg_{A}\) to the average distance between points of \(reg_{B}\) given in  by $q_r$. Overall resolution  \autoref{res} is the mean of $q_{r}$ over \(N\) regions. Resolution determines the level of detail in the point cloud. Low resolution can cause loss of texture and smaller objects making the point cloud unusable for applications that require high fidelity and detail.

\begin{equation}
    \qr{} = \frac{1}{N}\sum_{j=1}^N\left(\frac{\bar{dist}(reg_A^j)}{\bar{dist}(reg_B^j)}\right)
    \label{res}
\end{equation}
where, 
\vspace{-1cm}

\begin{gather*}
\bar{dist}(X) = \left(\frac{\sum \limits_{x^i \in X}{\min\limits_{x_j \in X} \lVert x_i - x_j \rVert_2}}{|X|}\right) ; i \neq j
\end{gather*}
    
\subsection{Accuracy \(\left(\qa{}\right)\)}
\label{Accuracy_Def}
We measure \emph{error} as the ratio of the sum of distances between every point in \(reg_{B}\) to the nearest neighbor in \(reg_{A}\) given distance is less than threshold \(\epsilon\) (shown in \autoref{pqm_animate}), to the product of the number of points in \(reg_B\) and \(\epsilon\), given by $q_{a}$, here \(\epsilon\) is the precision set by the user. The normalization is performed over (\(|reg_{B}|\times \epsilon\)) as this is the maximum distance possible if all points in \(reg_{B}\) are valid (i.e. have neighbors within \(\epsilon\) distance in \(reg_{A}\)). Accuracy is then measured as \(1 \ - \ \emph{error}\).  Overall accuracy ($\qa{}$) \autoref{accr} is the mean of $q_{a}$ over $N$ regions.

\vspace{-1em}
\begin{equation}
    \qa{} = \frac{1}{N}\sum_{j=1}^N \left(1\! -\! \left(\frac{1}{ \epsilon\, \lvert reg_{B}^j \rvert}\right)\! %
    \times\! %
    \sum\limits_{b \in reg_{B}^j} s(a, b) \right)%
    \label{accr}
\end{equation}

where, 
\vspace{-1em}
\begin{gather*}
    s(a,b)%
    =%
    \begin{cases}
      \min\limits_{a \in reg_A} \lVert a - b \rVert_2 & , \mathrm{if}\, \min\limits_{a \in reg_A} \lVert a - b \rVert_2 \le \epsilon \\
      0 & , \mathrm{otherwise}
    \end{cases}
\end{gather*}


Accuracy is computed on points that are not artifacts, which means any point not within set precision \(\epsilon\) is considered an artifact, and accuracy is not penalized for the same. This ensures that for a given change in the candidate, per point the change only contributes to either of the metrics. 

\subsection{Coverage \(\left(\qc{}\right)\)}
\label{Cov_Def}

Coverage is the ratio of  number of cells occupied by points of  \(pcd_A\) and \(pcd_B\) (shown in \autoref{pqm_animate}) to the number of cells occupied by points of \(pcd_A\). Coverage is computed per region as well as the whole point cloud separately, this gives us insights about local coverage (per cell) in addition to overall coverage. Overall coverage, unlike accuracy and resolution, is not an average of coverage per region over \(N\) regions, rather it is computed on the entire volume bounded by the two point clouds.

\begin{equation}
    Q_c = \left(\frac{|S_A \cap S_B|}{|S_A|}\right)
    \label{cov}
\end{equation}

\noindent Given by (\autoref{cov}) it is computed as a function of volume occupied in contrast to accuracy and resolution which are computed as functions of point-point distance. This ensures independence from accuracy and resolution sub-metrics, i.e. change in density of points or addition of Gaussian noise (within set precision \(\epsilon\)) has little to no effect on coverage, this is further explored in \autoref{eval_section}.

\subsection{Artifact Score \(\left(\qt{}\right)\)}
\label{Art_Def}

Artifacts are defined as the points in \(pcd_B\)  but not in \(pcd_A\) (shown in \autoref{pqm_animate}). These are generated due to reflections, distortion, or incorrect registration of points. Artifact score quantifies the lack of artifacts, i.e. the score is high if the candidate has low artifacts. We define artifacts as the ratio of number of cells occupied by points of \(pcd_B\) but not occupied by points of \(pcd_A\) to the number of cells occupied by points of \(pcd_B\). Artifact score is \(1 - Artifacts\). This, as shown in ~\autoref{art}, is similar to coverage in the way that it is computed as a function of the volume occupied as compared to point-point distance. 

\begin{equation}
    Q_t = 1 \ - \ \left(\frac{|S_B \setminus S_A|}{|S_B|}\right)
    \label{art}
\end{equation}

\subsection{Relationship between Metrics}
A challenge in identifying these metrics is to ensure that each of them are independent and that, together, they cover all aspects of map quality. Here, we analyze the independence of these metrics. 

{\bf Accuracy and Artifact-score}:
If points in \(pcd_B\) drift away from points in \(pcd_A\), \(\qa{}\) and \(\qt{}\) can both see change based on the \(\epsilon\) value set. If a point moves more than \(\epsilon\), it is counted as an artifact, affecting \(\qt{}\) but if it moves within the cell bounded by \(\epsilon\) it only affects \(\qa{}\) as defined in  \autoref{accr}. If noise is introduced to \(pcd_B\), both \(\qa{}\) and \(\qt{}\) can see change as some points may move out of the cells and some may move within.

{\bf Resolution and Accuracy}:
Since \(\qr{}\) and \(\qa{}\) are both distance-based metrics as shown in \autoref{res} and \autoref{accr}, they may appear to perform a similar role in quality measurement. However, \(\qr{}\) is measured with distances between points in the same point cloud, whereas \(\qa{}\) is measured with distances between points from different point clouds. In other words, \(\qr{}\) changes when points within \(pcd_B\) drift away from each other or when points within \(pcd_A\) drift away from each other, whereas \(\qa{}\) changes when points from \(pcd_B\) drift away from points in \(pcd_A\).

{\bf Resolution and Coverage}:
A change in \(\qr{}\) at lower values affects \(\qc{}\). When distances between points increase beyond \(\epsilon\), \(\qc{}\) decreases with \(\qr{}\) since there are gaps between points in the map. We note that this is consistent with the definition of \(\qc{}\). We also note that not all reductions in \(\qc{}\) will translate to a change in resolution. For example, a high-resolution map of a building floor with a room missing will be measured with lower coverage with no effect on \(\qr{}\).

Further, in \autoref{ablation_method}, we perturb the point cloud in various ways and show that \qq{} captures these perturbations in at least one of the metrics demonstrating comprehensiveness in quantifying map quality. 

\section{Evaluation}
\label{eval_section}

\begin{figure*}
  \centering
  \includegraphics[trim=1.2cm 1.2cm 1.2cm 1.2cm,clip=true,width=\textwidth]{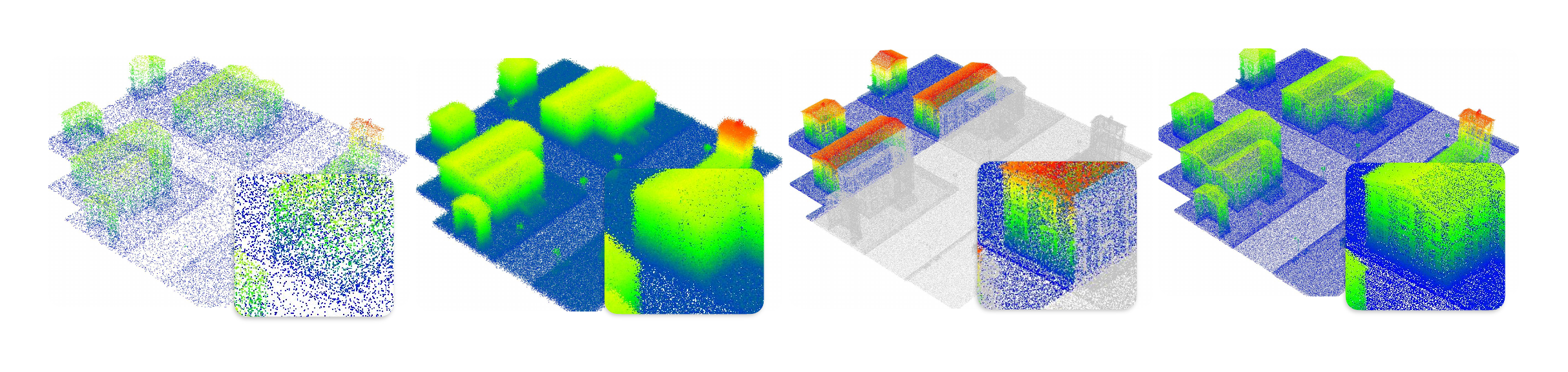}
  \caption{Ablation Study on \emph{street block} dataset; Left to right: Down-sampled, Noise Added, Cropped Simulated Artifacts}
  \label{unit_ablations}
  \vspace{-0.5cm}
\end{figure*}

We demonstrate the applicability of the proposed framework with extensive experimental evaluation. We start by performing an ablation study using a custom dataset which demonstrates \qq{}'s metrics' utility, independence and consistency in evaluating aspects of point clouds quality. Next, \qq{} is evaluated on two applications - dense SLAM and learning-based point cloud completion using simulation and real-world experiments. This demonstrates the broad applicability of \qq{} for a broad class of 3D perception applications and improves on other distance-based metrics.

\begin{table}
\caption{Ablation Study Results $\lvert\ \epsilon = 0.1$}
\label{unit_combined}
\resizebox{\columnwidth}{!}{%
\begin{tabular}{@{}c|c|cccccc@{}}
\toprule
Ablation & Value &  \cd{} & \hd{} &  $\qr{}$ & $\qa{}$ & $\qc{}$  & $\qt{}$ \\ 
\midrule

\multirow{2}{*}{Resolution}
&Uniform 75\%  & 504.91 & 0.17 & \textbf{0.87} & 1.00   & 0.93  & 1.00\\
&Uniform 50\%  & 1488.057 & 0.20 & \textbf{0.71} & 1.00   & 0.79  & 1.00\\

\midrule

\multirow{2}{*}{Accuracy}
&$\sigma=0.01$ &  682.95 & 0.55 & 0.91 & \textbf{0.84} & 0.92  & 0.77 \\
&$\sigma=0.02$ &  2064.36 & 0.1  & 0.81 & \textbf{0.73} & 0.93  & 0.80 \\
\midrule

\multirow{2}{*}{Coverage}  
& X, 40\% &  1.59E+08 & 24.27 & 1.00  & 1.00   & \textbf{0.40}  & 1.00   \\
& Y, 40\% &  1.38E+08  & 24.01 & 1.00 & 1.00   & \textbf{0.42} & 1.00  \\
\midrule

\multirow{2}{*}{Artifact} 
& X, +0.1m &  5846.90  & 0.1 & 0.99 & 0.62 & 0.79  & \textbf{0.79}\\
& XY, +0.282m & 2.63E+04 & 0.28 & 0.98 & 0.71 & 0.62  & \textbf{0.62}\\
\midrule

\end{tabular}%
}
\vspace{-0.7cm}
\end{table}

\subsection{Ablation Study}
\label{ablation_method}



%

The ablation study is performed using a prototype point cloud representing a simulated city-block bounded in a (40x40x10m) region containing approximately 1.28 million points (\autoref{unit_ablations}). We use simulation for accurate ground truth so we can study each metric of \qq{} in detail. The study involves applying various degradations to $pcd_A$ (source model), to produce a degraded model which is referred to as,  \(pcd_B\) in each case, and using \qq{}, \cd{} and \hd{}\footnote{\emd{} is not considered as heavy imbalance in candidate and reference point clouds prevents effective bijectivity / correspondence and \emd{} fails to provide any valuable insights into quality} to evaluate the quality at each step. \autoref{unit_combined} shows results of this experiment. 

\subsubsection{Resolution}
\label{ab_res}
We down-sample the source point clouds to simulate the reduction in resolution while preserving its overall structure. For each resolution ablation, we halve the total number of points. Uniform sampling is used to ensure points are removed consistently. When the resolution is reduced, the resolution metric $Q_{r}$ also decreases. Since the points are uniformly sampled, they create no artifacts in this study, which is reflected in the consistency of the $Q_{t}$ and $Q_{a}$ values.
An important observation is that \(Q_{c}\) changes with change in resolution. This is tied to the set precision. If a lower precision is set, the effect of the change in resolution is less on  \(Q_{c}\). 

\subsubsection{Accuracy}
 \label{ab_acc}
 To simulate loss of accuracy, Gaussian noise $\mathcal{N}(0,\,\sigma^{2})$ is applied to each axis of each point where $\sigma^2$ is the variance applied to the points in a random normal direction. This results in a significant change to each metric due to potentially shifting points into other regions, causing a loss of coverage, resolution, and an increase in artifacts. When a noise with $\sigma=0.01$ is applied, \(\qa{}\) shows a value of $0.8417$. If $\epsilon$ is increased to 0.2, the $Q_{a}$ score increases. This can be explained by how \(Q_{c}\), \(Q_{t}\) and \(Q_{r}\) are defined, adding Gaussian noise moves points out of cells into other cells affecting \(Q_c\) and \(Q_{t}\). Similarly, it also changes the distances between points as noise doesn't translate all points uniformly affecting \(Q_r\).

\subsubsection{Coverage}
\label{ab_comp}
Spatial coverage is reduced by cropping the point cloud along a particular axis to simulate a lack of coverage. \autoref{unit_combined} shows 2 ablated point clouds cropped to 40\% the original in both the X and Y axes. Results show that this is reflected in $Q_{c}$ as expected, due to its formulation as the ratio of the number of un-cropped points to the number of points in the original. 

\subsubsection{Artifacts}
\label{ab_art}
Artifacts are simulated by shifting the source point cloud resulting in points leaving their respective cells, thereby inducing artifacts. When a shift is applied in the X-axis, the resulting artifact score $Q_{t}$ drops down. However, if the precision ( $\epsilon$ ) is increased, the artifact score subsequently increases. This is because more points are considered valid when the $\epsilon$ value is increased. This same trend is observed when the point cloud is shifted in both X and Y with a larger value. This invariably affects coverage which is expected given points are non-uniformly distributed in the candidate.
This provides some insight into the effect of setting the precision, a higher precision (lower \(\epsilon\)) will lead to a higher artifact score.
\label{ablation_eval}

The ablation study shows that each our perturbations affect one of the \qq{} metrics but not others. For a real-world application, such comparisons provide hints on the benefits of using one method to construct 3D point clouds in comparison to another. Further, the ablation study shows that when precision is increased, the corresponding metric decreases noticeably. Precision is intended to be set based on expected quality and application and a very high precision i.e. \(\epsilon \rightarrow 0\) indicates a smaller expected margin of error. For example, when considering a robot manipulation application, the precision can be set based on the size of the objects being manipulated.

Overall, the experiment demonstrates the effectiveness of \qq{} in evaluating the quality of point clouds and detecting changes in quality due to different types of degradation. \qq{} metrics scale in a proportional manner with change to the point clouds and \(\epsilon\) provides control in quality assessment. This is in contrast to \cd{} and \hd{}'s behaviour where the change in these metrics cannot be effectively explained based on the degradation performed.

\begin{figure}[ht]
  \centering
  \includegraphics[trim=1.5cm 1.5cm 1cm 0.5cm,clip=true,width=\columnwidth]{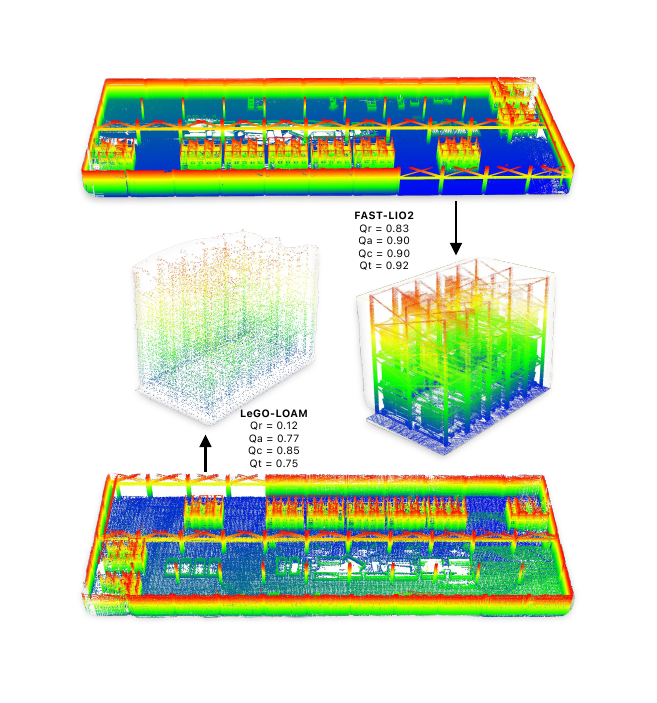}
  \caption{Simulation dataset; Point cloud built using FAST-LIO2 (Top) and LeGO-LOAM (Bottom). Zoomed in view for qualitative assessment}
\label{warehouse_demo}
  \vspace{-0.5cm}
\end{figure}

\begin{table}
\centering
\caption{Evaluation on SLAM maps  $\lvert\ \epsilon = 0.5$ \label{slam-table05} \charu{remove and refer main}}
\resizebox{\columnwidth}{!}{%
\begin{tabular}{c|c|cccccc}
\toprule
Dataset ($r$) & Method  & \cd{} & \hd{} & $\qr{}$ & $\qa{}$ & $\qc{}$ & $\qt{}$ \\ \midrule
\multicolumn{1}{l|}{\multirow{3}{*}{Davis (5)}}      & \multicolumn{1}{l|}{SHINE} & 2.13E+08         & \textbf{43.52}               & \textbf{0.95}       & 0.80     & 0.73     & 0.34           \\
\multicolumn{1}{l|}{}                                & \multicolumn{1}{l|}{LeGO}  & \textbf{3.67E+06}         & 79.20               & 0.28       & \textbf{0.86}     & 0.67     & \textbf{0.48}           \\
\multicolumn{1}{l|}{}                                & \multicolumn{1}{l|}{FAST}  & 4.79E+07         & 221.29              & 0.90       & 0.85     & \textbf{0.76}     & 0.40           \\ \midrule
\multicolumn{1}{l|}{\multirow{3}{*}{HILTI EX04 (2)}}      & \multicolumn{1}{l|}{SHINE} & 6.74E+06         & 6.63                & 0.80       & 0.73     & 0.74     & 0.30           \\
\multicolumn{1}{l|}{}                                & \multicolumn{1}{l|}{LeGO}  & 1.54E+06         & 6.46                & 0.12       & 0.75     & 0.66     & 0.26           \\
\multicolumn{1}{l|}{}                                & \multicolumn{1}{l|}{FAST}  & \textbf{6.02E+05}         & \textbf{5.56}                & \textbf{0.84}       & \textbf{0.84}     & \textbf{0.79}     & \textbf{0.62}           \\ \midrule
\multicolumn{1}{l|}{\multirow{3}{*}{Mai City (20)}}       & \multicolumn{1}{l|}{SHINE} & 6.57E+08         & 21.84               & \textbf{1.00}       & 0.79     & 0.31     & 0.62           \\
\multicolumn{1}{l|}{}                                & \multicolumn{1}{l|}{LeGO}  & 6.14E+08         & \textbf{21.74}               & 0.11       & \textbf{0.86}     & 0.26     & \textbf{0.81}           \\
\multicolumn{1}{l|}{}                                & \multicolumn{1}{l|}{FAST}  & \textbf{5.60E+08}         & 21.80               & 0.77       & 0.83     & \textbf{0.36}     & 0.73           \\ \midrule
\multicolumn{1}{l|}{\multirow{3}{*}{Warehouse (10)}} & \multicolumn{1}{l|}{SHINE} & 7.85E+06         & \textbf{3.96}                & \textbf{1.00}       & \textbf{0.85}     & \textbf{0.83}     & 0.81           \\
\multicolumn{1}{l|}{}                                & \multicolumn{1}{l|}{LeGO}  & 1.18E+07         & 12.05               & 0.12       & 0.71     & 0.62     & 0.62           \\
\multicolumn{1}{l|}{}                                & FAST                       & \textbf{7.39E+06}         & 11.87               & 0.83       & 0.82     & 0.78     & \textbf{0.85}           \\ \bottomrule
\end{tabular}%
}
\vspace{-0.5cm}
\end{table}

\begin{figure*}
  \centering
  \includegraphics[trim=1.5cm 1.5cm 1cm 0.5cm,clip=true,width=\textwidth]{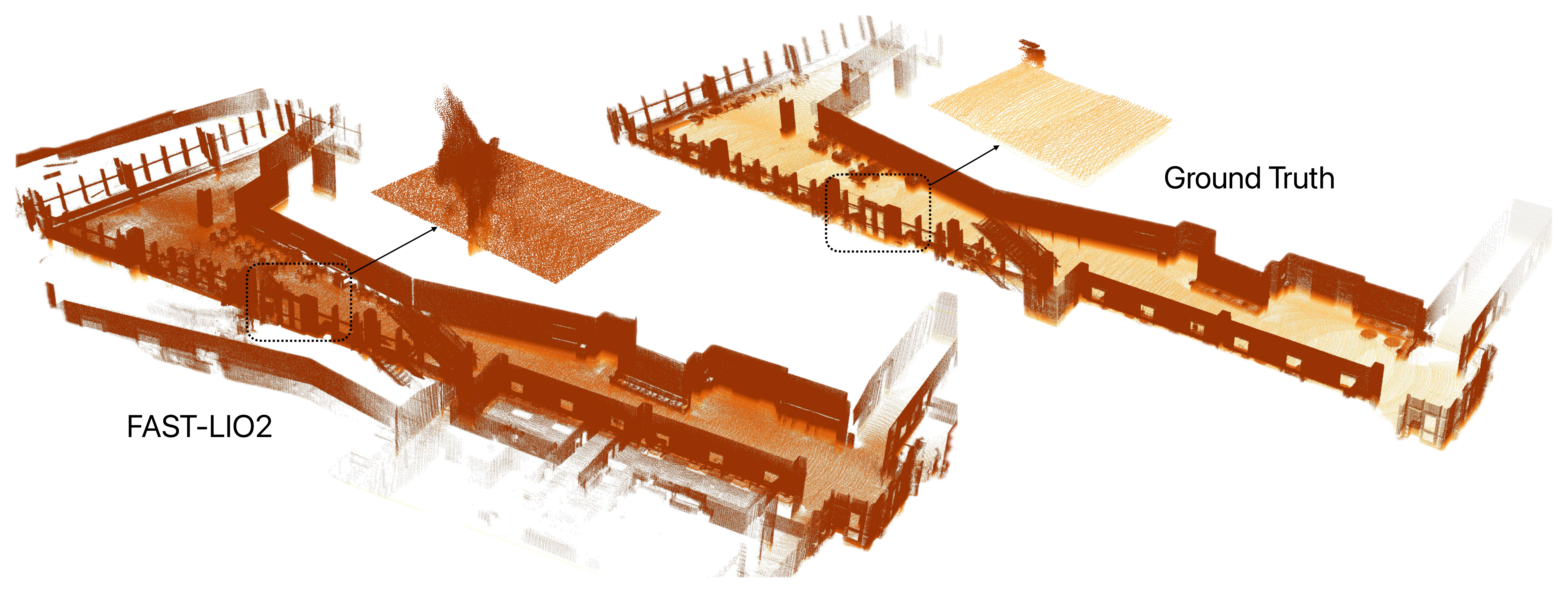}
  \caption{Real-world evaluation of Dense SLAM - Point clouds map generated using FAST-LIO2 (Spot robot + Ouster OS-1 128 LiDAR) on the left, and ground truth on the right ( robotic total-station).}
\label{davis_large}
\end{figure*}

\begin{figure}
  \centering
  \includegraphics[trim=1.5cm 1.5cm 1cm 0.5cm,clip=true,width=\columnwidth]{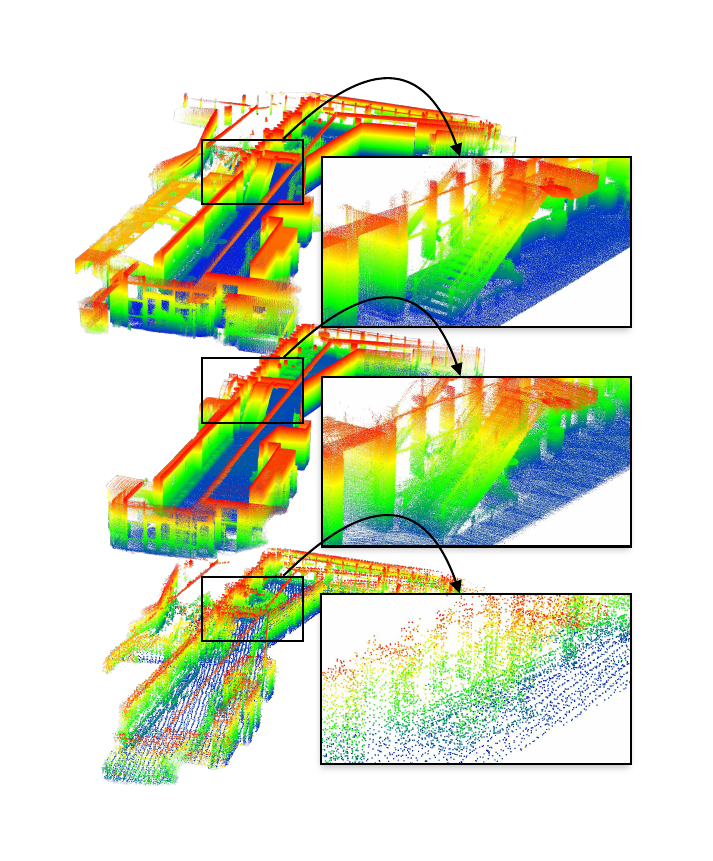}
  \caption{Evaluation on Davis dataset, zoomed-in view shows variations in detail for different SLAM methods. Top to Bottom: FAST-LIO2, Ground Truth, LeGO-LOAM. Zoomed in view of staircase on the right for qualitative assessment}
\label{davis_3map}
\end{figure}
\subsection{Evaluation on dense SLAM}
\label{slam_eval}
To further evaluate \qq{} we construct point clouds using dense SLAM methods. First, collect LiDAR scans in a custom simulation environment (\textit{citation removed for anonymity}) 
and build point cloud maps using several popular LiDAR SLAM systems. The simulation environments are built in Gazebo \cite{1389727} with mesh models of the worlds and a simulated Ouster OS1-128 LiDAR mounted on a Clearpath Husky \cite{clearpathroboticsHuskyOutdoor,githubGitHubHuskyhusky_simulator}. Worlds include \emph{Mai City} \cite{vizzoPoissonSurfaceReconstruction2021} and another named \emph{Warehouse} (\autoref{warehouse_demo}). These are designed to represent real-world environments with elements commonly found in the real world. Ground truth meshes are sampled to obtain ground truth point cloud since \qq{} compares point clouds. We also match the number of points from the maximum of the candidate point clouds to keep the comparison fair.

\begin{figure*}
  \centering
  \includegraphics[trim=1.5cm 1.5cm 1cm 0.5cm,clip=true,width=\textwidth]{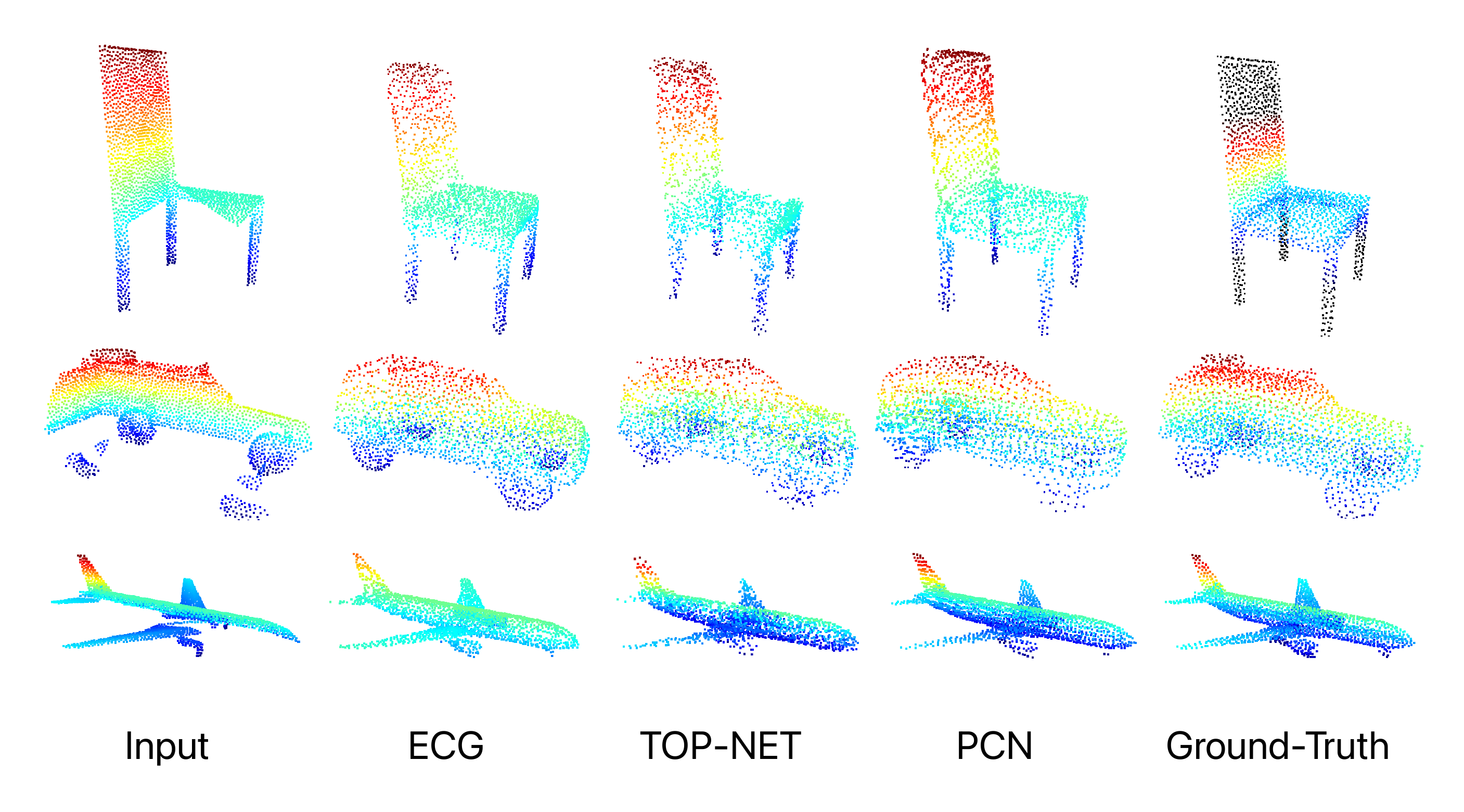}
  \caption{Point clouds generated using three completion networks; left to right: Input (partial cloud), ECG \cite{pan_ecg_2020}, TOP-NET \cite{tchapmi_topnet_2019}, PCN \cite{yuan_pcn_2018} and ground truth. Qualitative results are corroborated by quantitative evaluation with \qq{} shown in  \autoref{learning_table} }
\label{learning_demo}
\end{figure*}

Next, we test \qq{} on real-world data where collect LiDAR scans using an Ouster OS1-128 LiDAR and pose using a Leica Geosystems TS15 Robotic Total Station \cite{leicageosystemsRoboticTotal}. These scans are stitched using the captured poses and ICP \cite{rusinkiewiczEfficientVariantsICP2001,githubGitHubCloudCompareCloudCompare} to generate a ground truth map with \emph{mm precision} in poses. The dataset is named \emph{Davis} for ease of reference (\autoref{davis_large}, \autoref{davis_3map}). For the candidate point cloud, we capture LiDAR and IMU data using a Boston Dynamics Spot equipped with an Ouster OS1-128 LiDAR with a built-in IMU by walking it in the same building and use a visual SLAM method to build corresponding point cloud map. We also test \qq{} on \cite{zhangHiltiOxfordDatasetMillimetreAccurate2022} which contains a ground truth point cloud generated using an engineering-grade LiDAR. 
We study these four datasets (two simulation, two real-world) with three candidate SLAM methods totalling 12 point clouds with their corresponding ground truth.
Candidate point clouds are generated using LeGO-LOAM \cite{shanLeGOLOAMLightweightGroundOptimized2018}, FAST-LIO2 \cite{xufastlio2fastdirect2022}, and SHINE \cite{zhong2023icra}. The first two output a dense point cloud and real-time odometry while the last one employs a unique approach to mesh generation by employing hierarchical implicit neural representations to generate a mesh. \\
Note: The output mesh is sampled into a point cloud similar to the simulation ground truth, this leads to resolution metric \(Q_r = 1\) as the average distance between points is the same as the ground truth.

Finally, these are used to evaluate \qq{} metrics as well as \cd{} and \hd{}\footnote{\emd{} is not considered as heavy imbalance in candidate and reference point clouds prevents effective bijectivity / correspondence and \emd{} fails to provide any insights into quality} . The results of this evaluation are presented in \autoref{slam-table05}. 
\autoref{warehouse_demo} and \autoref{davis_3map} show the qualitative differences between the point clouds for the \emph{Davis} and \emph{Warehouse} datasets \footnote{Note: Evaluation is performed on the entire point cloud and not only the zoom-in view}. Each metric evaluates a certain aspect of quality as described in  \autoref{method}. We set precision $\epsilon = 0.5$ and $r$ based on point cloud size for all tests in  \autoref{slam-table05}. Other values were explored but not presented in the interest of space as results are consistent with the definition.\footnote{Comprehensive results provided in supplementary section along with figures}

Visually, it is apparent that the point cloud map generated with FAST-LIO2 has significantly higher detail than the one built using LeGO-LOAM. Subsequently, FAST-LIO2 receives higher \(Q_{r}\), \(Q_{a}\) and \(Q_{c}\) but a lower \(Q_{t}\) while \cd{} and \hd{} identify LeGO-LOAM as the one nearest to ground truth. In this example, \cd{} and \hd{} do not provide any insight into the point clouds' quality, i.e. the extreme imbalance between the point densities of the two candidates. It is vital to note that \qq{} metrics may not always agree with our perception of quality due to the multi-dimensional nature of quality assessment. However, they are fundamentally true to their definition which is consistent in \(\text{R}^3\) and accurate based on (\(\epsilon\)). We see this in point cloud maps built using LeGO-LOAM; although sparse, they are accurate. Their sparse nature contributes to the lack of artifacts which is reflected in \(Q_{t}\) but negatively affects \(Q_{c}\) and \(Q_{r}\).

\subsection{Evaluation of Learning-based Point Cloud Completion}

\begin{table}[]
\centering
\caption{Evaluation on Point cloud completion  \(\| \ \epsilon = 0.3\) \label{learning_table}}
\resizebox{\columnwidth}{!}{%
\begin{tabular}{@{}c|c|cccccc@{}}
\toprule
Model    & Method  & \cd{} & \hd{} & \(\qr{}\) & \(\qa{}\) & \(\qc{}\)     & \(\qt{}\)     \\ \midrule
\multirow{3}{*}{Airplane} & ECG     & 0.23                   & 0.03                   & 1         & 0.83                                     & \textbf{0.90} & 0.94          \\
                          & TOP-NET & 0.28                   & 0.03                   & 1         & 0.80                                     & 0.89          & 0.90          \\ 
                          & PCN     & \textbf{0.22}          & 0.03                   & 1         & \textbf{0.83}                            & 0.89          & \textbf{0.98} \\\midrule
\multirow{3}{*}{Car}      & ECG     & 1.17                   & 0.09                   & 1         & 0.58                                     & 0.61          & \textbf{0.78} \\
                          & TOP-NET & 1.50                   & 0.09                   & 1         & 0.57                                     & 0.56          & 0.67          \\ 
                          & PCN     & \textbf{1.22}          & \textbf{0.08}          & 1         & \textbf{0.58}                            & \textbf{0.65} & 0.75          \\\midrule
\multirow{3}{*}{Chair}    & ECG     & 0.85                   & \textbf{0.04}          & 1         & 0.61                                     & 0.71          & \textbf{0.80} \\
                          & TOP-NET & 1.08                   & 0.08                   & 1         & 0.63                                     & 0.69          & 0.69          \\ 
                          & PCN     & \textbf{0.72}          & 0.05                   & 1         & \textbf{0.66}                            & \textbf{0.82} & 0.73          \\\bottomrule
\end{tabular} 
}
\end{table}

A recent approach to point cloud generation is point cloud completion. We analyze three networks that output a completed point cloud when given an partial point cloud. For this study we consider  PCN~\cite{yuan_pcn_2018}, TopNet~\cite{tchapmi_topnet_2019} and ECG~\cite{pan_ecg_2020} point cloud completion models and the MVP dataset~\cite{pan_variational_2021} for their evaluation.
The resulting completed point clouds are evaluated against ground truth using \qq{}, \cd{} and \hd{}.  \autoref{learning_table} shows the quantitative results of this experiment while  \autoref{learning_demo} shows the resulting point clouds. On visual inspection of the point clouds, it is evident that the point clouds generated using ECG and PCN exhibit the highest quality and this is corroborated by \qq{}'s metrics. Unlike the SLAM experiments, these findings are corroborated by \cd{} and \hd{}. This reinforces our hypothesis regarding \cd{} and \hd{}'s limitations; both these metrics are able to identify the highest quality point clouds when the size of the pointcloud is small and where the density is roughly uniform but fail to do so in the SLAM study due to large size and unevenness of the point clouds.
Learning-based methods are rapidly becoming the dominant way to generate point clouds including point cloud completion~\cite{yuan_pcn_2018, tchapmi_topnet_2019, pan_ecg_2020}, image-based 3d reconstruction~\cite{fan2017point}, and image-to-mesh generation~\cite{Wang_2018_ECCV} etc., and we believe that \qq{} is the right framework to compare and evaluate their outputs.

\subsection{Compute Performance}
\label{compute_perf}

\begin{figure}
  \centering
  \includegraphics[trim=1.5cm 1.2cm 1.5cm 0cm,clip=true,width=\columnwidth]{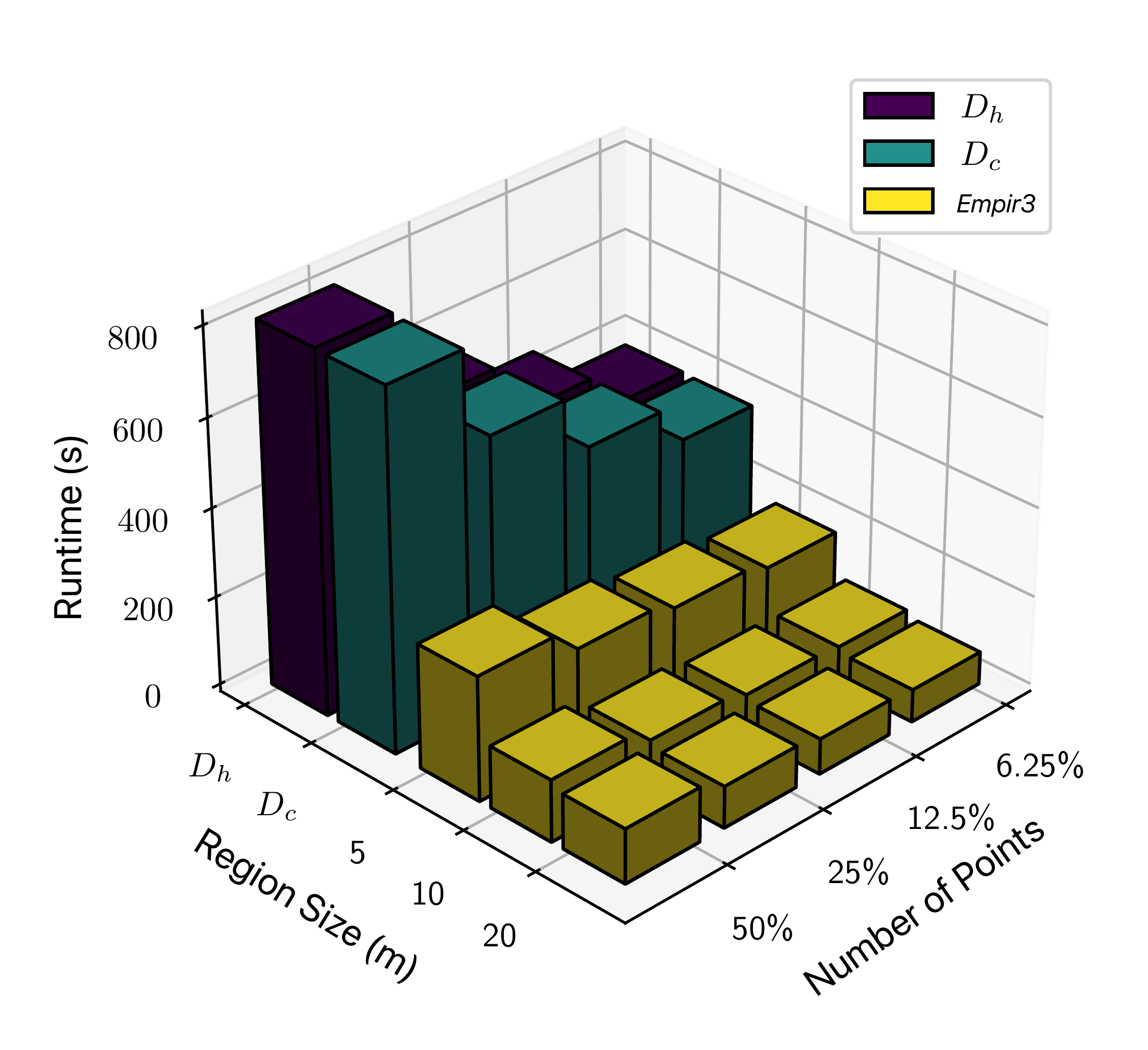}
  \caption{Plot shows runtime (Z-axis) of \qq{}, \cd{} and \hd{} on point clouds of varying resolution (Y-axis) and region sizes (X-axis). Candidate Map: Warehouse dataset with point cloud generated using FAST-LIO2 (Number of Points at 100\% = 67,690,672).}
  \label{compute-plot}
  \vspace{-0.5cm}
\end{figure}

\qq{} is implemented using Open3D \cite{Zhou2018}, PDAL \cite{pdalcontributorsPDALPointData2020}, Scikit-Learn \cite{scikit-learn}, NumPy \cite{harris2020array} and PyTorch \cite{pytorch} and is intended for dense point clouds ($> 10^7$ points).
 To handle such large point clouds, we provide a fast multi-threaded implementation that computes the regions in parallel, as well as a GPU-accelerated implementation capable of utilizing a GPU if one exists. In addition to that \qq{} computes in \(O(n\log n)\) which is similar if not faster than popular methods while being multi-dimensional (\autoref{compute-plot}).
This experiment is conducted to compare our implementation of \qq{} with Chamfer and Hausdorff distances for 4 resolutions of the point cloud generated using FAST-LIO2 on the Warehouse dataset. Different resolution maps were obtained by down-sampling the original point cloud. As expected, reducing the number of points decreases the computation time for all. However, at different region sizes, we are at least twice as fast as \cd{} and \hd{}.

\section{Applications : Anomaly Detection}

\begin{figure}
  \centering
  \includegraphics[trim=1cm 1cm 1cm 1cm, clip=true,width=\columnwidth]{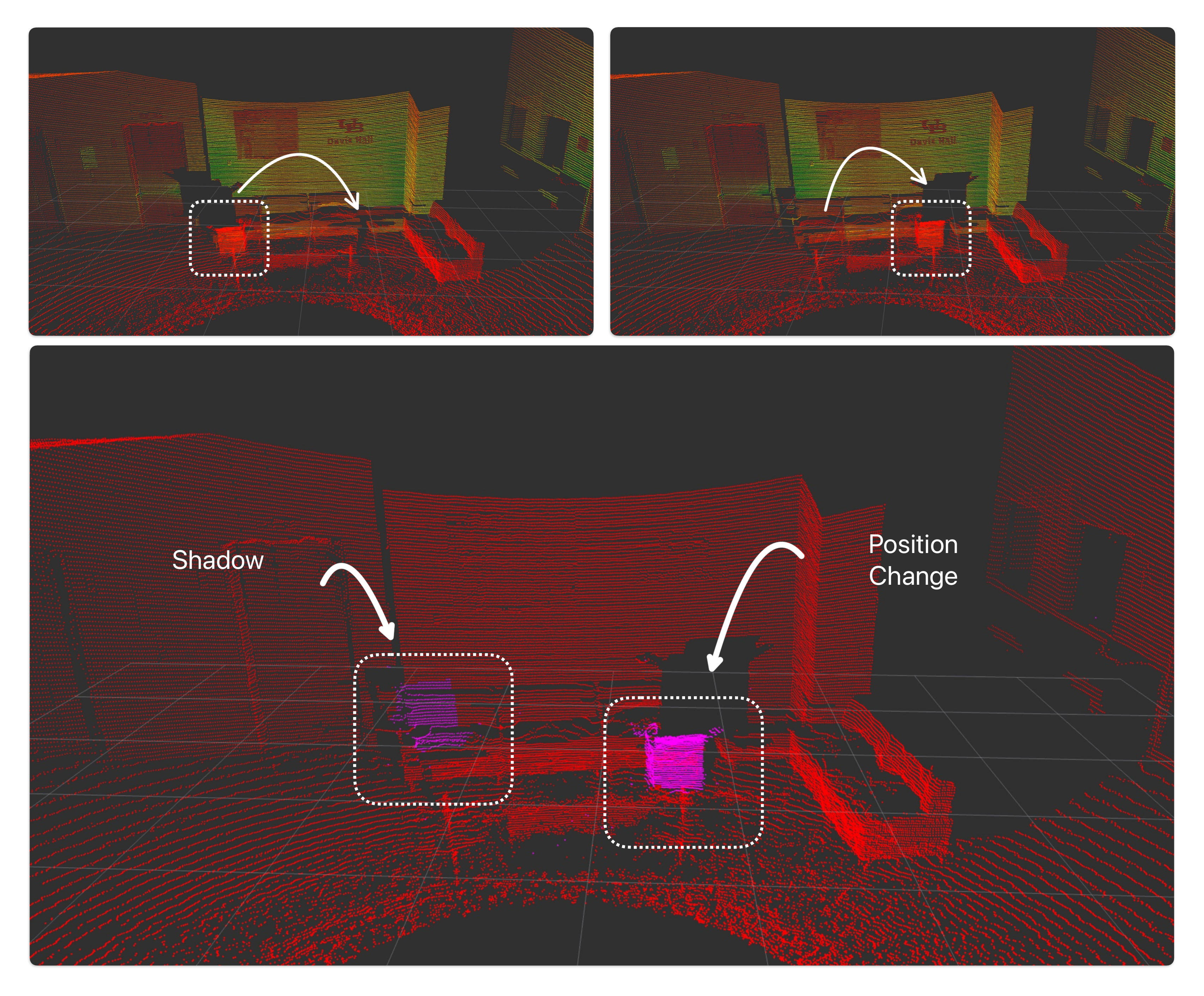}
  \caption{Top: Anomaly (change) detection using \qq{}. \qq{} allows real-time anomaly detection at $> 5$ FPS on an Ouster OS-1 128. Figure shows anomaly detection on sample dataset, a box is moved and anomalies are highlighted in purple. The anomalies measure out to 0.021 which indicates 2.1\% of the scene has changed.D}
\label{e3d-change}
  \vspace{-0.8cm}
\end{figure}


To explore additional applications of \qq{}, we propose its use for anomaly detection. The objective in anomaly detection is to measure and localize changes in a scene using a reference point cloud map. Traditional distance measures can achieve this but are often slow due to the reasons outlined in \autoref{related}. By leveraging \qq{}, we can accelerate this process, as \qq{} is computationally efficient and enables real-time performance.

To validate this, we conducted experiments using an Ouster OS-1 128 Channel LiDAR mounted on a Boston Dynamics Spot robot. The robot operated in a controlled environment where objects could be moved, added, or removed. \autoref{e3d-change} demonstrates results from one such experiment. In this scenario, a large cardboard box was relocated, and \qq{}‘s anomaly detection output is highlighted in purple. The detection identifies two regions of anomalies: new points (artifacts) are visible in the area where the box was originally placed, and the box’s new location shows coverage. The terms ‘coverage’ and ‘artifacts’ are interchangeable, depending on whether the initial LiDAR frame is considered the reference or the candidate. This interchangeability does not impact usability or performance.

Furthermore, we can quantify the detected change. The change illustrated in \autoref{e3d-change} measures to 0.021, indicating that 2.1\% of the scene has changed. We achieved real-time performance at 5 frames per second with a LiDAR output of 5.2 million points per second. Performance can be further enhanced by limiting the field of view (FOV) to the region of interest (ROI).
\section{Discussion}
\label{discussion}

{\bf Multi-Dimensional Evaluation}:
Evaluations in \autoref{eval_section} help demonstrate the utility of \qq{}'s multi-dimensional approach to quality assessment. We emphasize that the objective of \qq{} is not to categorically align with any specific qualitative assessment, but rather to illustrate and quantify various aspects of point quality. Such a comprehensive assessment provides valuable feedback to point cloud construction methods and allows the developers to improve them. It also allows specific applications to identify quantifiable metrics in point cloud quality and how they correspond to application accuracy (object recognition, for example). The comprehensive assessement is clearly articulated in the SLAM and point cloud completion experiments, where \cd{} and \hd{} only provide a single number while \qq{} is able to quantify each aspect of quality.


The assessments further highlight certain trends in the behaviour of \cd{} and \hd{}. Both these distance measures perform relatively well when the point clouds are small and densities are consistent. This is demonstrated in the evaluation of point cloud completion networks where they identify ECG and PCN's outputs as highest quality which is in agreement with \qq{} metrics and qualitative results. This behaviour does not hold when point clouds have high-imbalance and/or are large in size. In the SLAM experiments, in some datasets they identify LeGO-LOAM's point clouds as the ones with the highest quality which contradicts qualitative results. Overall, \cd{} and \hd{} fail to provide any real insight into the point clouds' quality.

{\bf Applications}:
Although \qq{} is demonstrated on dense point clouds for the scope of this paper, we anticipate applications in various domains such as 
\begin{itemize}
    \item Optimizing point cloud construction: Algorithms such as Visual SLAM intend to recreate 3D structure for navigation, manipulation etc. \qq{} provides better insight into the algorithm performance, thereby better informing the developer of how it could be used for the end application. 
    \item Improving learning on point clouds: Chamfer loss \cite{fan2017point} \cite{ravi2020pytorch3d} is a popular loss function used for 3D deep learning tasks. The various \qq{} metrics can provide a way to learn in a structured manner for applications such as depth completion, point cloud generation, etc.
    \item Sensor characterization: \qq{} can be used to quantify how well a sensor or suite of sensors are able to see all obstacles in a scene. Such characterization could be useful for a sensor suite on an autonomous car, for example, to identify potential blind spots.
\end{itemize}

{\bf Need for reference point clouds}: 
As \qq{} and other methods evaluated in this paper are full-reference similarity measures, ground truth point clouds are necessary for evaluation. Ground truth point clouds can be generated using better sensors (e.g., an engineering-grade LiDARs or Total-Station). Alternatively, we can evaluate methods in a simulation where ground truth is readily available. However, most point cloud construction methods need to be evaluated using a reference - and \qq{} requires the same. 

{\bf Limitations}:
We identify some limitations of \qq{} as a quality metric. \qq{} is not a distance measure, so doesn't scale the way Chamfer distance does. In its current state, \qq{} needs significant updates to be used as a distance measure. The four metrics identified for quality only apply to the point cloud's 3-dimensional information (x, y, z) and not other modalities such as intensity, reflectivity, and color information. 
\qq{} is also not differentiable which prevents it from being used directly as a loss function in its current form. Despite this constraint, \qq{} can be used to gain valuable understanding of learning-methods' capabilities and performance.

\section{Conclusion}
We propose \qq{}, a multi-dimensional point-cloud quality evaluation framework comprising coverage, artifact score, accuracy, and resolution metrics which can serve as a comprehensive tool to evaluate large point-clouds. \qq{} is designed to capture aspects of point cloud quality not addressed by existing evaluation metrics. Through detailed evaluations on four datasets, three SLAM algorithms and three learning-based point completion methods, we demonstrate the superiority of \qq{} in comparison to popular metrics such as \cd{} and \hd{}. 
\qq{} can be used to understand and assess the performance of point cloud construction algorithms such as Visual SLAM systems, learning-based approaches to point cloud completion, depth estimation and others. We conjecture that insights provided by \qq{} can also help developers better train their algorithms for these tasks either to improve overall quality or task-specific performance. We expect to open-source the \qq{} framework and associated simulation setup for broader use on publication of this work. 


\bibliographystyle{plainnat}
\bibliography{filtered,additional}

\end{document}